\documentclass[10pt,onecolumn,letterpaper]{article}

\usepackage{times}
\usepackage{graphicx}
\usepackage{amsmath}
\usepackage{amssymb}
\usepackage{subfigure}
\usepackage{float}
\usepackage[]{hyperref}

\begin{document}

\title{On Residual Networks Learning a Perturbation from Identity}
\date{}
\author{Michael Hauser$^1$\\
Children's Hospital of Philadelphia\\
{\small mikebenh@gmail.com}
}
\footnotetext[1]{Michael Hauser has been supported by a postdoctoral fellowship at the Center for Autism Research in the Children's Hospital of Philadelphia. Any opinions, findings and conclusions or recommendations expressed in this publication are those of the author and do not necessarily reflect the views of the sponsoring agencies.}
\maketitle

\begin{abstract}
The purpose of this work is to test and study the hypothesis that residual networks are learning a perturbation from identity. Residual networks are enormously important deep learning models, with many theories attempting to explain how they function; learning a perturbation from identity is one such theory. In order to answer this question, the magnitudes of the perturbations are measured in both an absolute sense as well as in a scaled sense, with each form having its relative benefits and drawbacks. Additionally, a stopping rule is developed that can be used to decide the depth of the residual network based on the average perturbation magnitude being less than a given epsilon. With this analysis a better understanding of how residual networks process and transform data from input to output is formed. Parallel experiments are conducted on MNIST as well as CIFAR10 for various sized residual networks with between 6 and 300 residual blocks. It is found that, in this setting, the average scaled perturbation magnitude is roughly inversely proportional to increasing the number of residual blocks, and from this it follows that for sufficiently large residual networks, they are learning a perturbation from identity.
\end{abstract}

\section{Introduction}

Residual networks were designed from the intuition that it is easier to learn a mapping that is a perturbation from the identity than it is to learn an unreferenced map~\cite{he2016deep}. The purpose of this work is to test and study whether this intuition that guided the design of the residual network is in fact the means by which they process data. This is important since many of the state-of-the-art machine learning algorithms are set by neural networks that use some type of skip connection~\cite{xie2017aggregated}, and so understanding how these networks function is important to both help guide the design of networks of these types, as well as intersting from a purely theoretical standpoint. It should be noted that Densely Connected Convolution Networks~\cite{huang2017densely} have skip connections, but they were designed from the intuition that they allow features to be preserved as they flow through the network, as opposed to learning a perturbation from identity.

\subsection{Common Understandings of Residual Networks}

Three of the most popular ways in which residual networks are understood to operate will be briefly discussed. The first is as they were originally designed, as a perturbation from identity~\cite{he2016deep}. This line of thinking lends itself to the interpretation that they are finite difference approximations to differential equations~\cite{hauser2017principles}, so in fact the residual network is learning a differential equation mapping input to output. This follows because a perturbation from identity is exactly a first order Taylor expansion, where the perturbation is the forcing function over a given time step.

The second is that residual networks, with their identity skip connections, can be reorganized via simple algebraic invariances to the collection of all possible forward paths of normal non-skip connection neural networks~\cite{veit2016residual}. This means that residual networks are equivalent to ensembles of non-skip connection neural networks, and when trained in this more explicit way, the classification performance was shown to increase.

The third is that residual networks, with a single hidden node per layer and ReLU activation, is a universal approximator as the depth of the network goes to infinity~\cite{lin2018resnet}. (This is in contrast to Kolmogorov's universal approximation theorem, where the width of the network goes to infinity). A rough idea of the mechanism by which the residual network with ReLU activation can do this is that it is learning something analogous to a simplicial complex over the data manifold, although the complex is by no means a collection of $d$-dimensional triangles, but instead whatever piecewise linear surfaces the network learns.

There is much value in studying each of these perspectives, and under different settings each of them can become a more or less accurate model for understanding how the network transforms the data under consideration. This work however will concern itself only with the first mode of understanding, namely that the residual network is learning a perturbation from identity. This work also studies a standard residual network architecture~\cite{he2016identity} on two standard computer vision tasks (MNIST and CIFAR10), which suggests that, at least in these common settings, this is how residual networks function.

\section{Depth and Computational Complexity}

This section develops the model of residual networks based from perturbation theory. In the first subsection we will develop the understanding directly from the common understanding of residual networks, namely that of an identity skip connection added to the nonlinear forcing function. In the second subsection we will draw analogies directly from systems of ordinary differential equations.

In this paper we will refer to a residual block in the same sense as was used in the original work on residual networks~\cite{he2016deep}, namely as the combination of the identity term plus the nonlinear forcing function. Similarly, we will refer to a residual section as the region of the residual network where the image size stays the same. When the height and width are halved by doubling the stride and the number of channels is doubled by doubling the number of channels, there is a new residual section.

\subsection{ResNets as Perturbations from Identity}
\label{sec:math-section}

Residual networks have the layerwise update of the following form, where \\ $l \in \left\lbrace 0,1,2,\dots,L-1 \right\rbrace $ for a section of the residual network with $L$-many residual blocks per section:

\begin{equation}
\label{eqn:resnet}
    x^{(l+1)} = x^{(l)} + f \left( x^{(l)} ; l \right)
\end{equation}

In order to test the presupposition that Equation~\ref{eqn:resnet} is in fact a mapping that is a perturbation from identity, we want $ || f \left( x^{(l)} ; l \right)||_2 << ||x^{(l)}||_2 $, or equivalently:

\begin{equation}
\label{eqn:perturbation-inequality}
    \frac{|| f \left( x^{(l)} ; l \right) ||_2}{||x^{(l)}||_2} =
    \frac{|| x^{(l+1)} - x^{(l)} ||_2}{||x^{(l)}||_2}
    << 1
\end{equation}

We use this measure because the magnitude of the first order perturbation term should be much less than the magnitude of the zero order term.

Additionally, this line of thinking allows us to directly measure the \emph{computational distance}~\cite{hauser2019state} the data travels through the network. The distance travelled between residual blocks $l$ and $l+1$ is given by:

\begin{equation}
\label{eqn:discrete-trajectory}
    || x^{(l+1)} - x^{(l)} ||_2 = || f \left( x^{(l)} ; l \right) ||_2
\end{equation}

This means that the total distance the data travels in a residual section is given by:

\begin{equation}
\label{eqn:total-distance}
    d := 
    \sum_{l'=0}^{L-1} || x^{(l'+1)} - x^{(l')} ||_2 = 
    \sum_{l'=0}^{L-1} || f \left( x^{(l')} ; l' \right) ||_2
\end{equation}

Explicitly we are measuring the magnitude of the perturbation in a scaled, normalized sense according to $ \frac{|| x^{(l+1)} - x^{(l)} ||_2}{||x^{(l)}||_2} $, as well as in an absolute sense according to $ || x^{(l+1)} - x^{(l)} ||_2 $. The first is important for quantifying to what extent residual networks are learning a perturbation from identity, while the second is important for quantifying how much, as measured by distance, the residual network is transforming the data, as in Equation~\ref{eqn:total-distance}.

For $N$-many images of size height$\times$width$\times$channels, the scaled size of a perturbation at layer $l$ is given as the average of $ \frac{|| x^{(l+1)} - x^{(l)} ||_2}{||x^{(l)}||_2} $ over the $N$-many images. Similarly for the absolute difference of $ || x^{(l+1)} - x^{(l)} ||_2 $, this value is take as the average over the $N$-many images.

\subsection{Analogy to Continuous Layer Systems}

Equation~\ref{eqn:resnet} is the forward difference approximation to a first order system of ordinary differential equations, where $ l \in \left[ 0,1 \right]$ is now a continuous variable indexing the \emph{continuous} depth of the neural network:

\begin{equation}
\label{eqn:first-order-ode}
\frac{dx^{(l)}}{dl} = f \left(x^{(l)};l \right) 
\end{equation}

To see this, doing a first order Taylor expansion of $x^{(l+\Delta l)}$ around $x^{(l)}$ yields the approximation $x^{(l+\Delta l)} \approx x^{(l)} + \frac{dx^{(l)}}{dl} \Delta l $, and inserting Equation~\ref{eqn:first-order-ode} into this yields:

\begin{equation}
\label{eqn:first-order-expansion}
    x^{(l+\Delta l )} 
    \approx 
    x^{(l)} + f \left(x^{(l)};l  \right) \Delta l 
\end{equation}

This assumes a partitioning of the layerwise variable $l\in \left[0,1\right]$ where $l=0$ is the "start" layer and $l=1$ is the "end" layer, in our case we go from a continuous transformation in the layers to a discrete one. The partitioning of the continuous layers is the following: 

\begin{equation}
\label{eqn:partitioning}
    \mathcal{P} := \\
    \left\lbrace 
    0=l\left(0\right) < l\left(1\right)<\dots<l\left(n\right)<\dots<
    l\left(L-1\right) = 1
    \right\rbrace
\end{equation}

Defining $ \Delta l := l\left(n+1\right) - l\left(n\right)  $, this partitioning is valid for approximating the solution of Equation~\ref{eqn:first-order-ode} by Equation~\ref{eqn:first-order-expansion} as long as  $ \max_n \Delta l \rightarrow 0 $ as $ L\rightarrow \infty $. Note that $\Delta l$ is a function of $n$ so we could explicitly write it as $ \Delta l \left(n\right)$, but in order to keep the notation simpler we only write $ \Delta l $.

Additionally, notice that the continuous time analogue to the velocity of the data through the (discrete) network, namely Equation~\ref{eqn:discrete-trajectory}, is given by the magnitude of the (continuous) derivative:

\begin{equation}
    || \frac{dx^{(l)}}{dl} ||_2  =  || f \left(x^{(l)};l \right) ||_2 
\end{equation}

Again in analogy to Equation~\ref{eqn:total-distance}, the total distance the image particle travels through the network is given by integrating over the entire residual section:

\begin{equation}
    d :=
    \int_{0}^{1} || \frac{dx^{(l')}}{dl} ||_2 dl' = \\
    \int_{0}^{1} || f \left(x^{(l')};l' \right) ||_2 dl'
\end{equation}

This is the continuous layer analogy of distance the data particle travels to the actual discrete layer increments of the residual network.

\subsection{Stopping Rule for Network Depth}

The perturbation term of Equation~\ref{eqn:first-order-expansion}, namely $\Delta l$, is \emph{implicit} in the neural network's activation function. By this we mean there is no $\Delta l$ written \emph{explicitly} in Equation~\ref{eqn:resnet}, and so to measure the magnitude of the perturbation for the residual network we will test the presupposition of Equation~\ref{eqn:perturbation-inequality} by \emph{defining} this to be the magnitude of the perturbation, namely:

\begin{equation}
    \Delta l :=
    \frac{|| x^{(n+1)} - x^{(n)} ||_2}{||x^{(n)}||_2}
\end{equation}

The neural network is therefore \emph{learning} the partitioning of Equation~\ref{eqn:partitioning} as it learns to map input to output. It is not learning the total number of layers per residual section $L$, as this is obviously a user defined parameter, and it is not completely learning the distance the data travels $d$ as this value seems to be dependent on the data set and to a certain extent independent of the total number of layers. Instead it is learning where each of the partition regions $l\left(n\right)$ for $n \in \left\lbrace 0,1,\dots ,L-1 \right\rbrace$ are placed.

If this picture of residual networks is correct, we can expect that the average size of $\Delta l$ over the $L$-many layers, namely $ \Bar{\Delta l} = \frac{1}{L} \sum_{n=0}^{L-1}\Delta l $ to go down as $\frac{1}{L}$. In fact, if there is a fixed \emph{computational distance} the data travels, denoted as $ d_{\text{scaled}}$, then the average size of the perturbations should go as follows:

\begin{equation}
\label{eqn:delta-bar}
    \bar{\Delta l} = \frac{d_{\text{scaled}}}{L}
\end{equation}

Furthermore this line of thinking immediately suggests a systematic way for defining the number of layers for the neural network. Specifically once the distance $d_{\text{scaled}}$ is estimated, for any $\epsilon > 0 $ one can then take $L = \text{ceil} \left( \frac{d_{\text{scaled}}}{\epsilon}\right) $ to ensure that $ \bar{\Delta l} < \epsilon $, where $ \text{ceil} \left( \cdot \right)$ is the ceiling function that rounds the argument up to the next integer.

\subsection{Distances as a Metric Geometry}

Data manifolds have intrinsic shapes associated with them. From a technical perspective this means that there is a metric tensor associated with the data manifold, where the metric tensor is used to measure distances on the surface of the data manifold. In Equation~\ref{eqn:total-distance} we are measuring distances, not along the spatial surface of the data manifold, but instead in the layerwise direction. The closed form solution for the metric tensor on the surface of the data manifold, as learned by the neural network, can thus be extended to be able to measure distances along the combined curved space-layer manifold.

\begin{equation}
    ds^2 = dx^{(l) \hspace{1 pt} 2} + g \left( x^{(l)} \right)_{a_lb_l} 
    dx^{a_l} 
    dx^{b_l}
\end{equation}

Here we have defined $ dx^{(l) \hspace{1 pt} 2} :=  || f \left( x^{(l')} ; l' \right) ||_2^2 $ from Equation~\ref{eqn:discrete-trajectory} while the metric tensor $ g \left( x^{(l)} \right)_{a_lb_l} $ on the surface of the data manifold comes from~\cite{hauser2017principles}. Compared to General Relativity, the layerwise dimension is analogous to the time dimension, the metric is positive definite instead of Lorenztian, and here all of the off-diagonal terms are zero so the layerwise dimension is decoupled from the spatial dimensions.

\section{Experimental Results}

These hypotheses are tested in the experimental section of this paper. Namely, experiments on both CIFAR10 as well as MNIST were performed. Both residual networks had the pre-activation form~\cite{he2016identity} of Equation~\ref{eqn:resnet}, with the nonlinear forcing function as follows:

\begin{equation}
f \left( x^{(l)} ; l \right) := \\
    W_2^{(l)}* \text{ReLU} \left( \text{BN}_2^{(l)} \left(
    W_1^{(l)}* \text{ReLU} \left( \text{BN}_1^{(l)} \left(x^{(l)}\right) \right) \right) \right)
\end{equation}
where $\text{BN}$, $\text{ReLU}$ and $W*$ are the batch-normalization, ReLU-activation and convolution operations, respectively. The same basic architecture was used for both experiments, where there were $3$ residual sections and $L$-many residual blocks per section. Custom implementations were written in the Python package TensorFlow~\cite{abadi2016tensorflow} and trained via error backpropagation~\cite{rumelhart1985learning}.

\begin{figure}[H] 
  \centering
  \subfigure[Test errors on MNIST]{\includegraphics[width=0.45\linewidth]{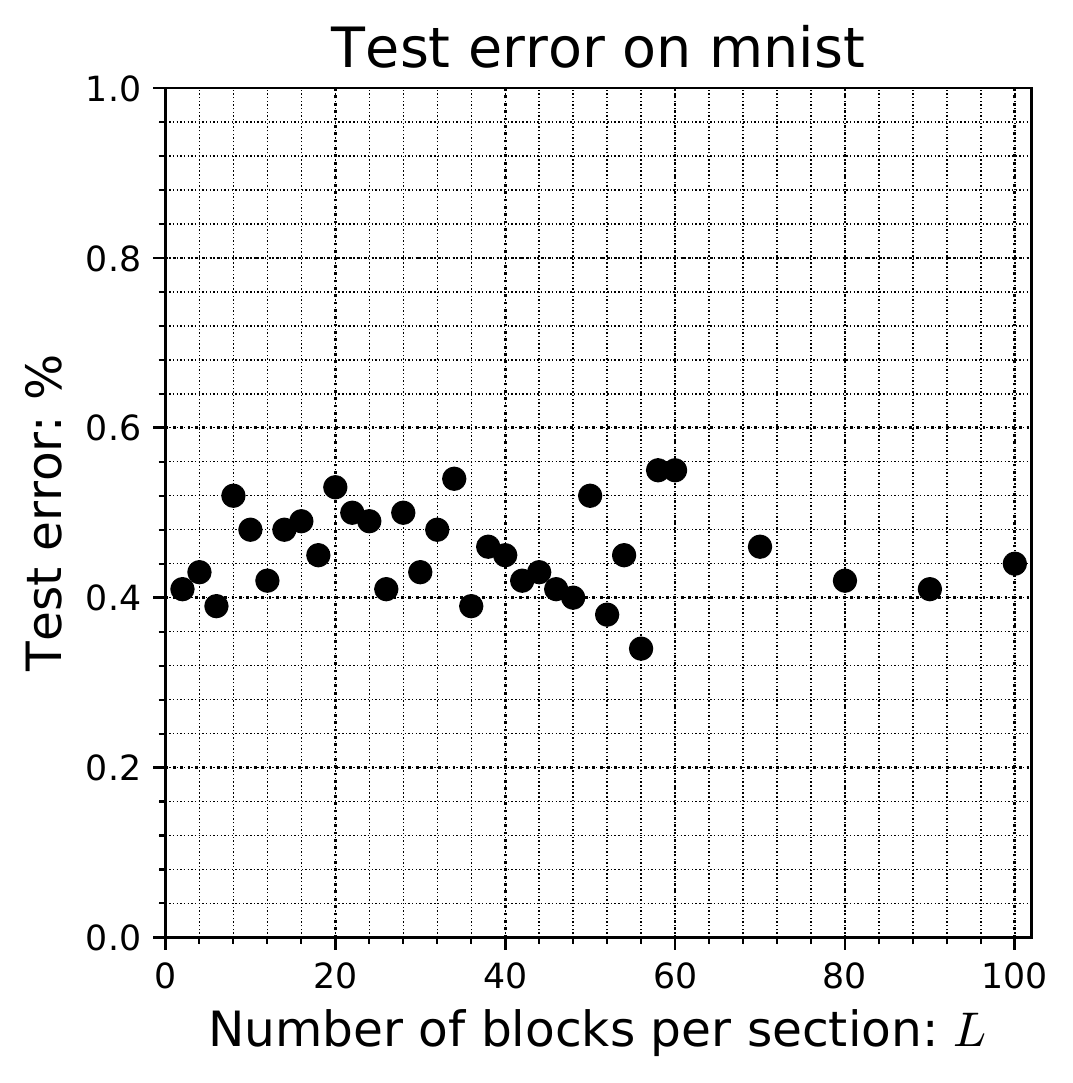}}\quad
  \subfigure[Test errors on CIFAR10]{\includegraphics[width=0.45\linewidth]{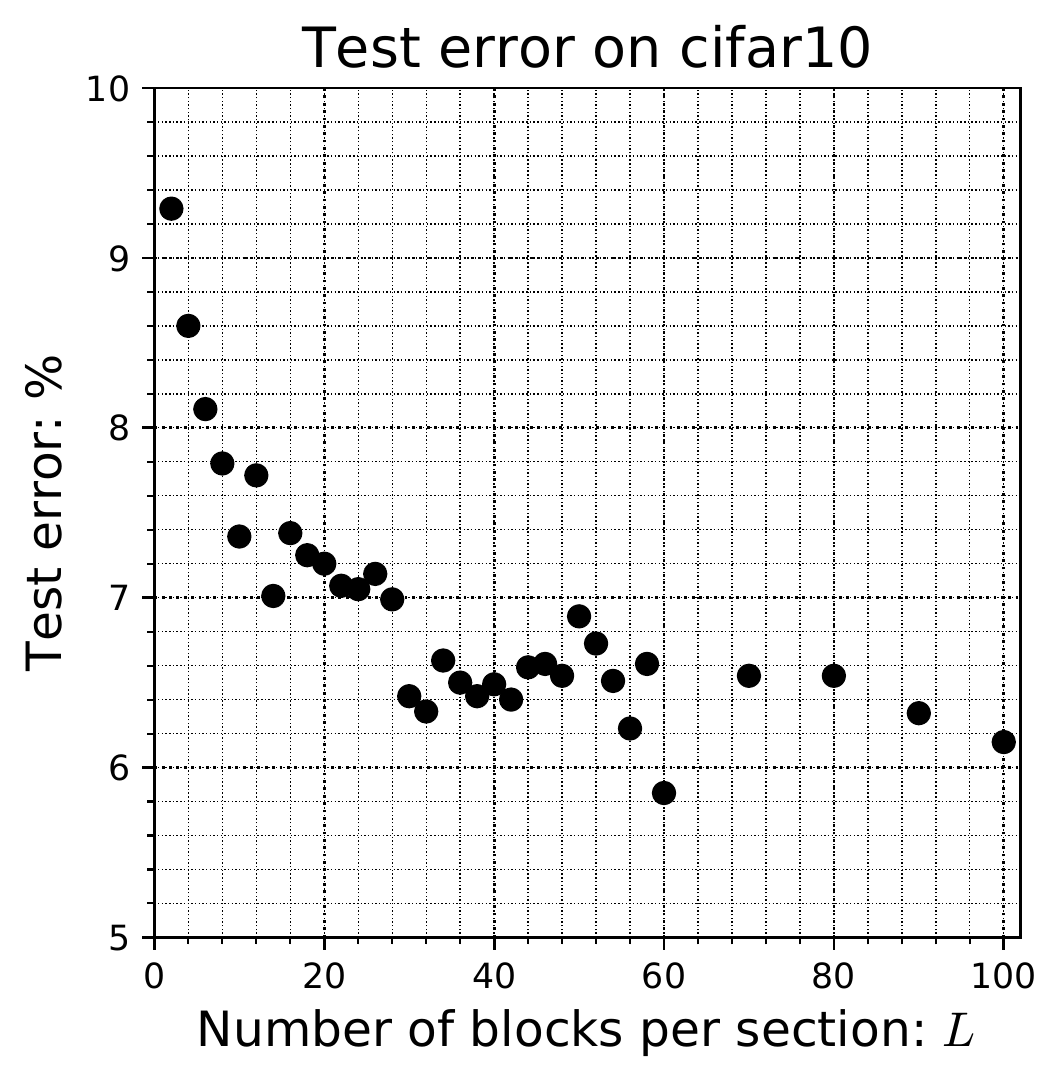}}
  \caption{Test errors on both MNIST as well as CIFAR10, with varying numbers of residual blocks per section. It is seen that the accuracies are more or less equal to reported implementations.}
\label{fig:accuracies}
\end{figure}

\begin{figure*}[t] 
  \centering
  \subfigure[Magnitude of perturbation between each block on MNIST.]{\includegraphics[width=0.95\textwidth]{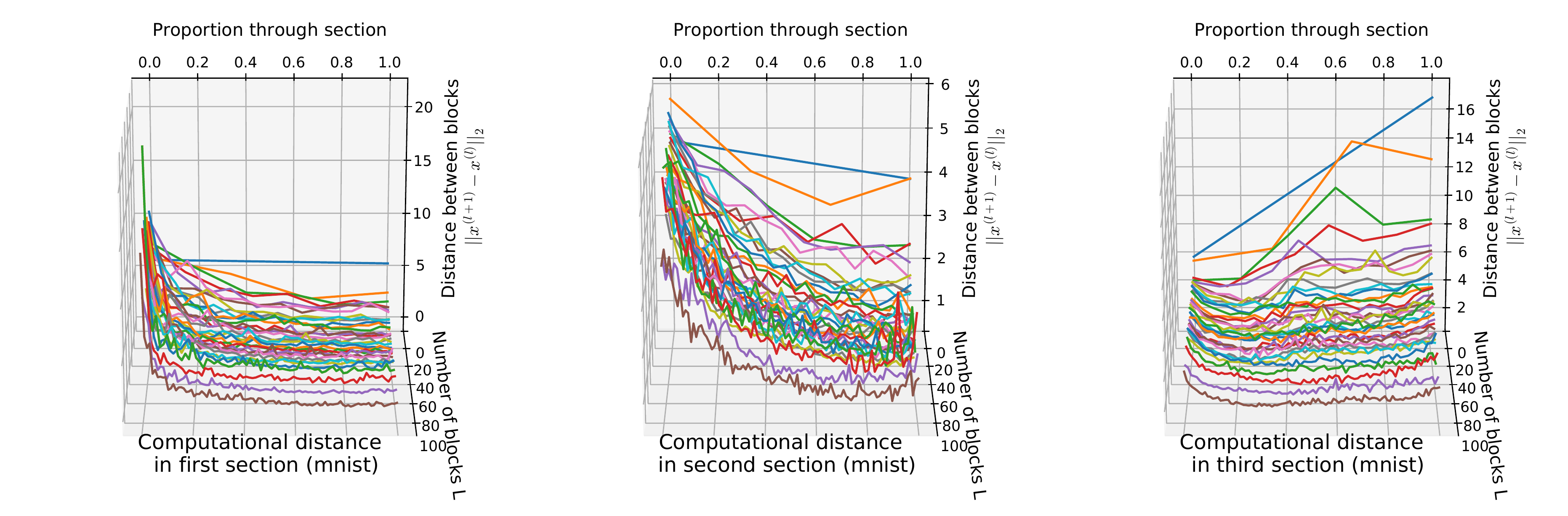}}\quad
  \label{fig:3dmnist}
  \subfigure[Magnitude of perturbation between each block on CIFAR10.]{\includegraphics[width=0.95\textwidth]{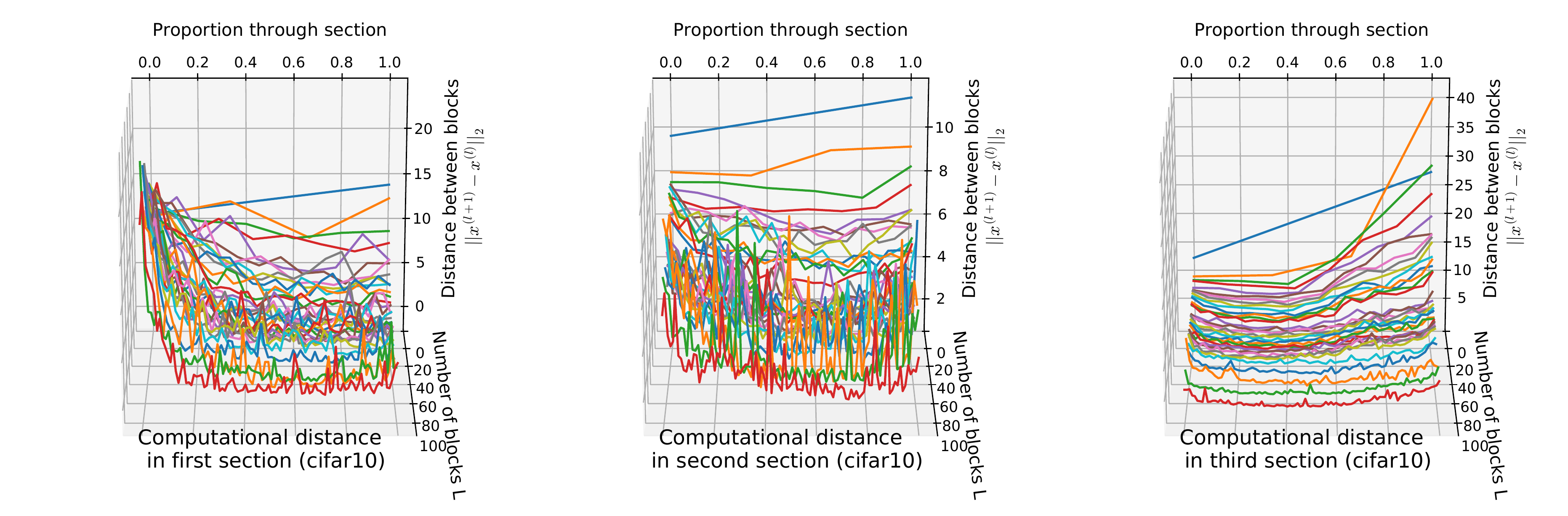}}
  \label{fig:3dcifar10}
  \caption{Magnitudes of the perturbations for each block in the network. The axis from left to right denotes the distance (scaled from $0$ to $1$) through the residual section, so for example the first blue line in each of the six figures is for $L=2$ and so it has two points constructing the line, and hence it is straight (the colors exist only to help distinguish between the different experiments). The axis out of the page is for the various sized networks we trained, with the number of residual blocks ranging over $L=2,4,\dots,68,70,80,90,100 $. Finally the vertical axis is the dependent variable measuring the magnitude of the perturbation $ ||x^{(l+1)}-x^{(l)}||_2 $, and thus is it measuring if a given block is making a larger or smaller transformation of the data.}
\label{fig:3d}
\end{figure*}

The first, second and third sections operated on images (denoted by height $\times$ width $\times$ channels) of sizes $32\times32\times16$, $16\times16\times32$, $8\times8\times64$ for CIFAR10 and $28\times28\times16$, $14\times14\times32$, $7\times7\times64$ for MNIST. To change between sections, $1\times1$ filters were used to double the number of channels, while a stride of $2$ was used to halve the height and width. The only image augmentations used (only on CIFAR10) were random horizontal flips, and random cropping after the image was padded with $4$-many $0$'s on each side.

The test errors of the models, with varying numbers of residual blocks-per-section, can be seen in Figures~\ref{fig:accuracies}. This is to show that the models we used achieve test errors at rates more or less equal to official implementations. Because of memory limitations, on CIFAR10 we trained the models for $L=2,4,\dots,28$ with batch sizes of $256$, $L=30,32,\dots,58$ with batch sizes of $128$ and finally $L=60,70,\dots,100$ with batch sizes of $64$. Similarly for MNIST, we trained the models for $L=2,4,\dots,60$ with batch sizes of 128, while for $L=70,80,90,100 $ we used batch sizes of 64. Note that when converted to layers, with $3$ residual sections, $L$-many blocks per section and $2$ layers-per-block, our networks have $6\cdot L+2$ layers; the $2$ come from the input to expand to $16$ channels as well as the fully connected layer at the output after global average pooling.

\subsection{Magnitudes of the Perturbations}

\begin{figure*}[t] 
  \centering
  \subfigure[Computational distances on MNIST.]{\includegraphics[width=0.95\textwidth]{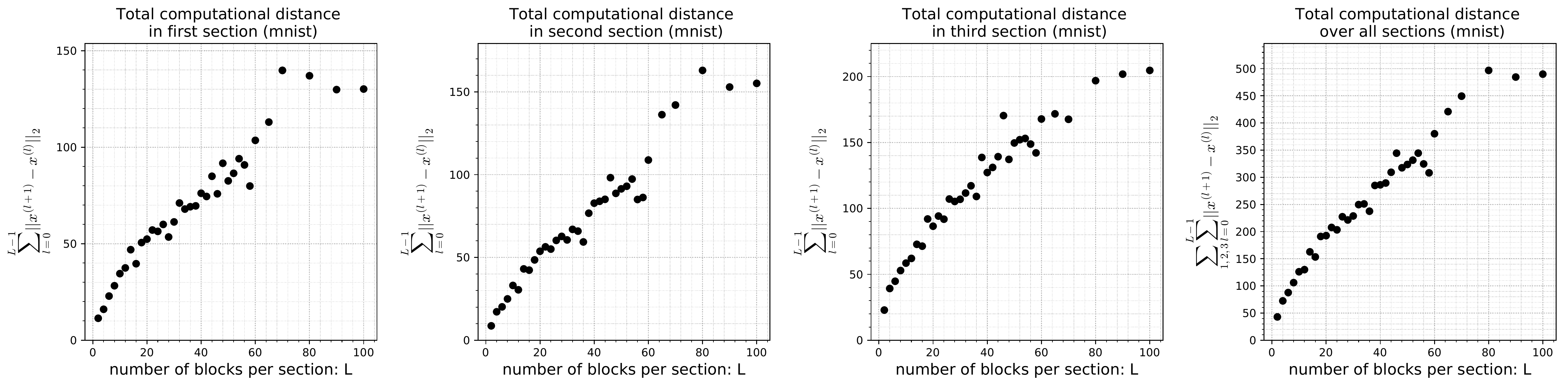}}\quad
  \subfigure[Computational distances on CIFAR10.]{\includegraphics[width=0.95\textwidth]{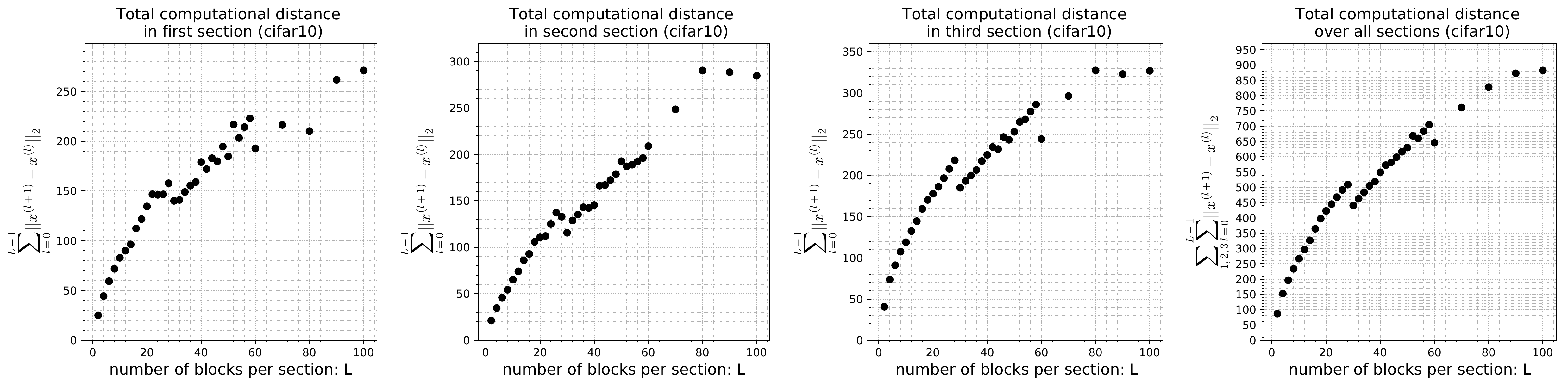}}
  \caption{Computational distances in the three residual sections of the network, namely $ \sum_{l'=0}^{L-1} ||x^{(l+1)}-x^{(l)}||_2 $ for $L=2,4,\dots,68,70,80,90,100$, as well as the sum of the the three sections to produce the total computational distance, namely $ \sum_{\text{section}=1,2,3} \sum_{l'=0}^{L-1} ||x^{(l+1)}-x^{(l)}||_2 $.}
\label{fig:mnist-distances}
\end{figure*}

\begin{figure*}[t] 
  \centering
  \subfigure[Mean perturbation magnitudes on MNIST.]{\includegraphics[width=0.95\textwidth]{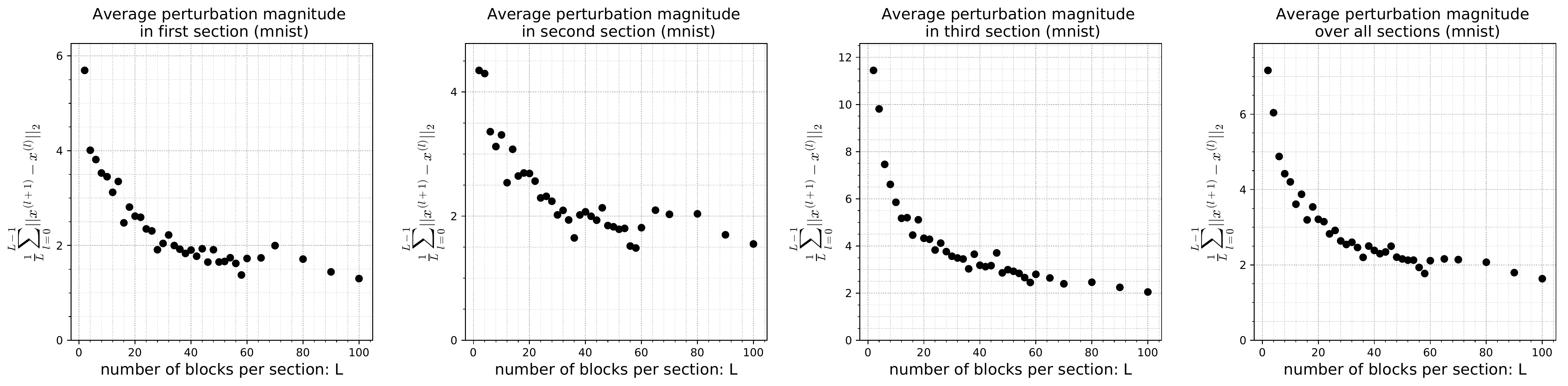}}\quad
  \subfigure[Mean perturbation magnitudes on CIFAR10.]{\includegraphics[width=0.95\textwidth]{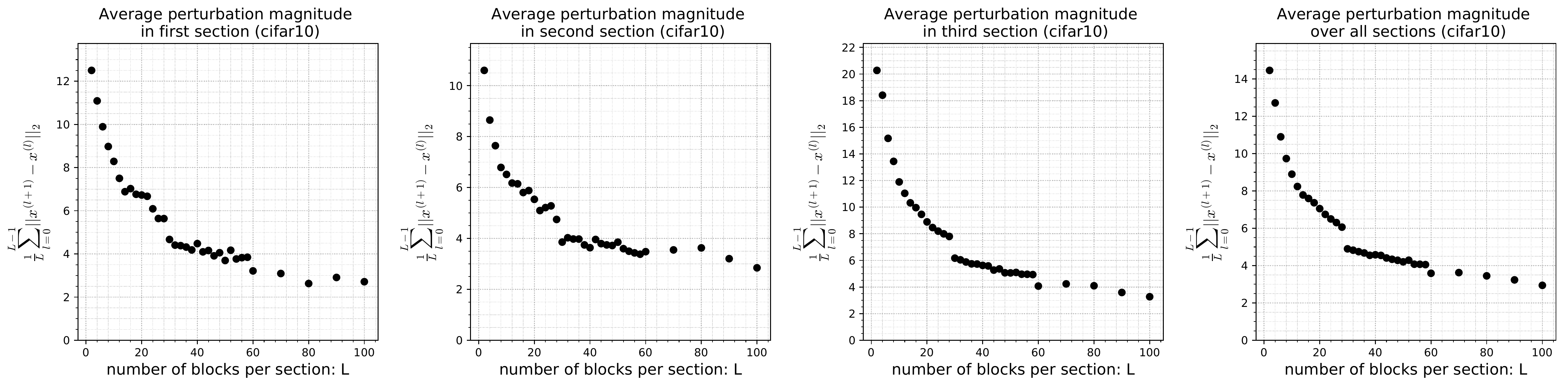}}
  \caption{Mean perturbation magnitudes in the three residual sections of the network, namely $\frac{1}{L} \sum_{l'=0}^{L-1} ||x^{(l+1)}-x^{(l)}||_2 $ for $L=2,4,\dots,68,70,80,90,100$, as well as the mean perturbation magnitude over the entire network, namely $ \frac{1}{3\cdot L}\sum_{\text{section}=1,2,3} \sum_{l'=0}^{L-1} ||x^{(l+1)}-x^{(l)}||_2 $.}
\label{fig:avg-distance-size}
\end{figure*}

This subsection examines the magnitude of the perturbations as the data passes through the network, and can be seen in Figure~\ref{fig:3d}. Specifically, this is measuring the magnitude of $ || x^{(l+1)} - x^{(l)} ||_2 $ individually for each $l \in \left\lbrace 0,1,\dots,L-1\right\rbrace$ for each of the three residual sections.

Figures~\ref{fig:3d}a and~\ref{fig:3d}b show the computational distances the MNIST and CIFAR10 images travel through each residual block of the network, respectively. It is seen that in the first and second residual sections, the earlier blocks in the section more significantly transform the image than the later blocks of the same residual section. This seems to suggest that the earlier blocks of the residual sections make larger, more coarse transformations to the image, while later blocks make smaller, more fine grained transformations. This could explain why the stochastic depth residual networks~\cite{huang2016deep} tend to work better when the earlier blocks are dropped out at lower rates than the later blocks. 

In the third residual section however, both the earlier and later blocks contribute more, while the middle blocks contribute less. We speculate that this could be a consequence of backpropagation more significantly changing the blocks towards the end of the network than the beginning, since backpropagation starts at the output and works towards the input.

\subsection{Computational Distance}
\label{sec:measuring-computational-distance}

\begin{figure*}[t] 
  \centering
  \subfigure[Magnitude of scaled perturbation between each block on MNIST.]{\includegraphics[width=0.95\textwidth]{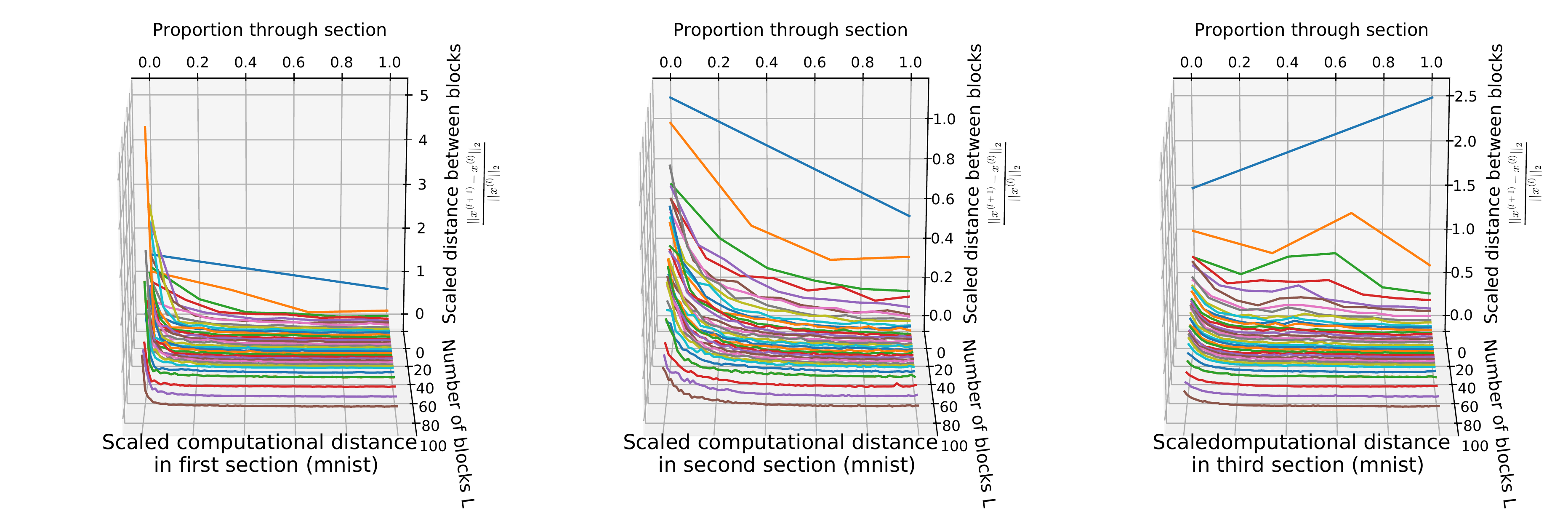}}\quad
  \subfigure[Magnitude of scaled perturbation between each block on CIFAR10.]{\includegraphics[width=0.95\textwidth]{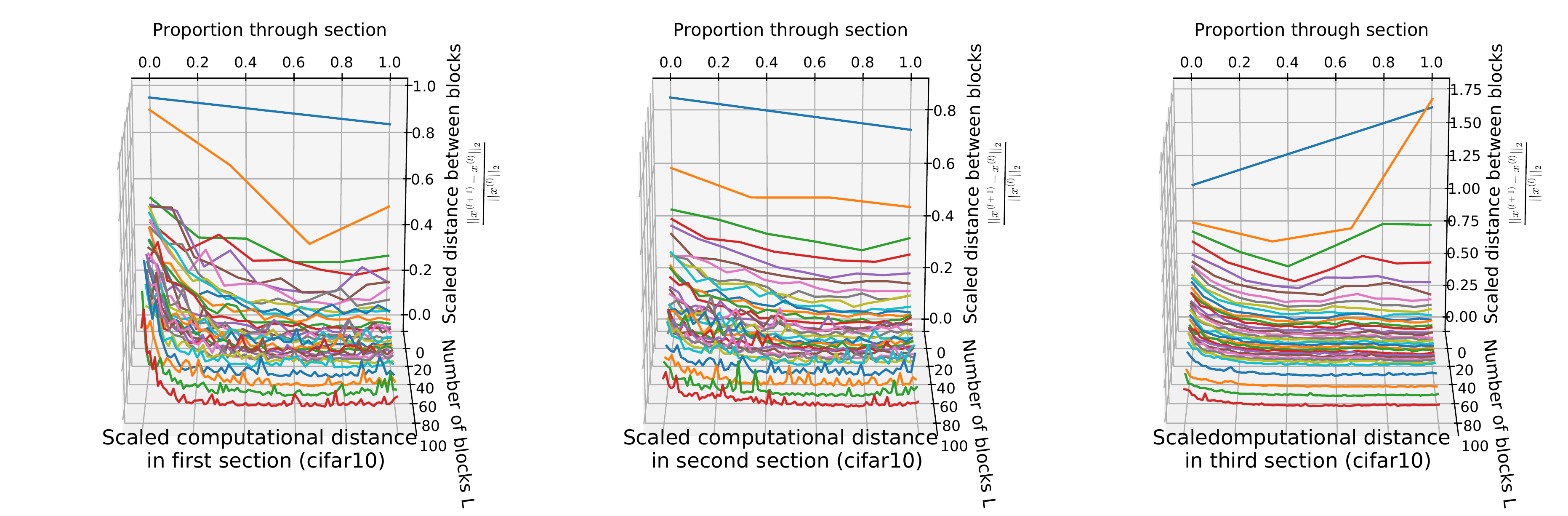}}
  \caption{Magnitudes of the scaled perturbations for each block in the network. The axis from left to right denotes the distance (scaled from $0$ to $1$) through the residual section, so for example the first blue line in each of the six figures is for $L=2$ and so it has two points constructing the line, and hence it is straight (the colors exist only to help distinguish between the different experiments). The axis out of the page is for the various sized networks we trained, with the number of residual blocks ranging $L=2,4,\dots,68,70,80,90,100 $. Finally the vertical axis is the dependent variable measuring the magnitude of the scaled perturbation $ \frac{||x^{(l+1)}-x^{(l)}||_2}{||x^{(l)}||_2} $, and thus is it measuring if a given block is making a larger or smaller transformation of the data.}
\label{fig:3d-scaled}
\end{figure*}

This subsection examines the computational distance the data travels through the network, with varying numbers of residual blocks-per-section, according to Equation~\ref{eqn:total-distance}, namely $ d := \sum_{l'=0}^{L-1} || x^{(l'+1)} - x^{(l')} ||_2 $. In the continuous sense, this can be thought of as integrating each of the velocity curves of Figure~\ref{fig:3d} to yield the total distance the particle travels, for varying numbers residual blocks $L$.

Figure~\ref{fig:mnist-distances}a shows the computational distances the MNIST images need to travel from input to output, while Figure~\ref{fig:mnist-distances}b shows the computational distances the CIFAR10 images need to travel. Several conclusions are drawn from these results.

First it is seen that the computational distance tends to increases as the number of blocks per section increases. This makes sense as one would expect that the more blocks used to transform the images, the more transformed the images become. It is interesting to note however that there is a diminishing return on depth in terms of transforming the images. By this we mean for smaller networks (roughly $L<20$ for MNIST), as one doubles the number of blocks, the total distance the data travels tends to about double. However for more medium sized networks (roughly $20<L<80$ for MNIST) as one doubles the number of blocks, the total distance the data travels increases, but by an amount less than half. Finally for the large networks (roughly $L>80$ MNIST) there seems to be \emph{no} increase in computational distance when increasing the number of blocks, which brings us to the second conclusion.

Second it is seen that the computational distance the data travels for each of the three residual sections increases up until about $L=80$ blocks per section, which suggests that, for this type of network, forcing function, dataset (as well as other training parameters), it takes about $L=80$ finite difference regions to well-estimate the differential equation for each residual section. Notice that for $L=80$ residual blocks, this neural network has $482$ layers of linear operation and nonlinear activation, so this is a fairly large network. It is seen that, at this point for roughly $L>80$, the computational distance the MNIST images need to travel as they are transformed from low-level input to high-level output is 130, 160 and 200 for the first, second and third residual sections, leading to a total computational distance of 490, for the network to construct a high level representation of the image.

Similarly for the networks trained on CIFAR10, it is seen that for the smaller networks (roughly $L<15$) there is a doubling of computational distance corresponding to a doubling of the number of residual blocks. The more medium sized networks (roughly $15<L<70$) have a diminishing return of computational transformations for increasing the number of residual blocks in the network. Finally the larger networks (roughly $L>70$) tends to saturate and there seems to be \emph{no} increase in computational distance with increasing the number of residual blocks. At this point the computational distances of the first, second and third residual sections are 250, 290 and 320, yielding a total computational distance of 860 for the network to construct a high level representation of the image from a given low level input representation.

It is interesting to note that the computational distance for CIFAR10 is larger than the computational distance for MNIST. This makes sense as CIFAR10 is a more complex dataset than MNIST, and so one would expect that it would take more transformations to process CIFAR10 than it does to process MNIST.

\subsection{Average Perturbation Magnitude}

\begin{figure*}[t] 
  \centering
  \subfigure[Scaled computational distance on MNIST.]{\includegraphics[width=0.95\textwidth]{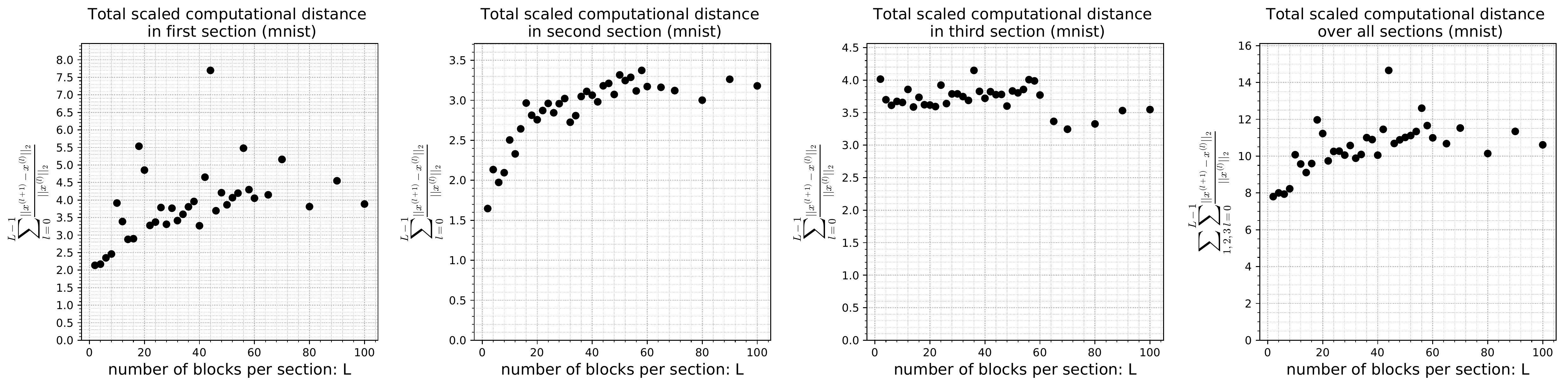}}\quad
  \subfigure[Scaled computational distance on CIFAR10.]{\includegraphics[width=0.95\textwidth]{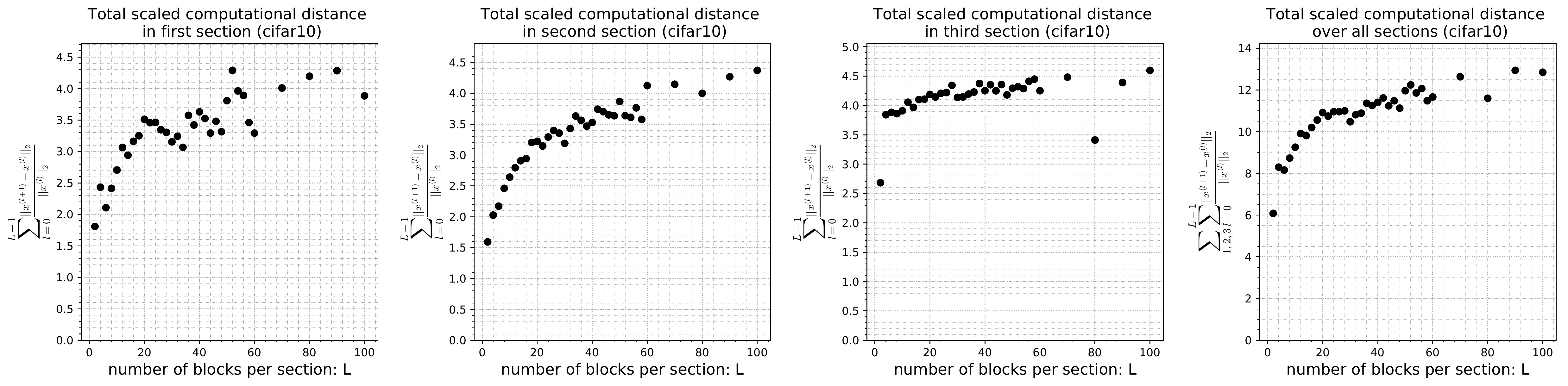}}
  \caption{Scaled computational distances in the three residual sections of the network, namely $ \sum_{l'=0}^{L-1} \frac{||x^{(l+1)}-x^{(l)}||_2}{||x^{(l)}||_2} $ for $L=2,4,\dots,68,70,80,90,100$, as well as the sum of the the three sections to produce the total scaled computational distance, namely $ \sum_{\text{section}=1,2,3} \sum_{l'=0}^{L-1} \frac{||x^{(l+1)}-x^{(l)}||_2}{||x^{(l)}||_2} $.}
\label{fig:scaled-comp-dist}
\end{figure*}

\begin{figure*}[t] 
  \centering
  \subfigure[Mean perturbation on MNIST.]{\includegraphics[width=0.95\textwidth]{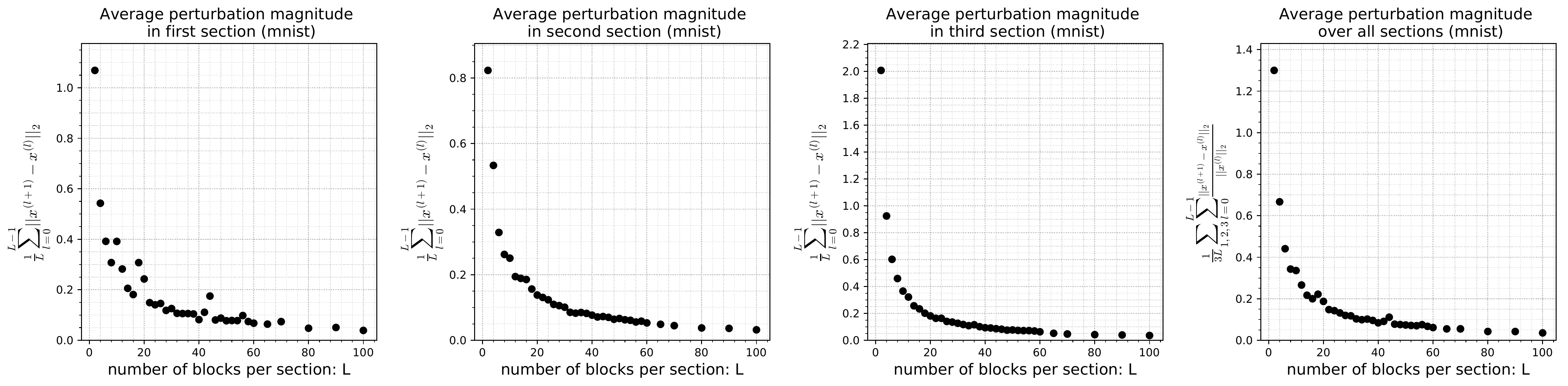}}\quad
  \subfigure[Mean perturbation on CIFAR10.]{\includegraphics[width=0.95\textwidth]{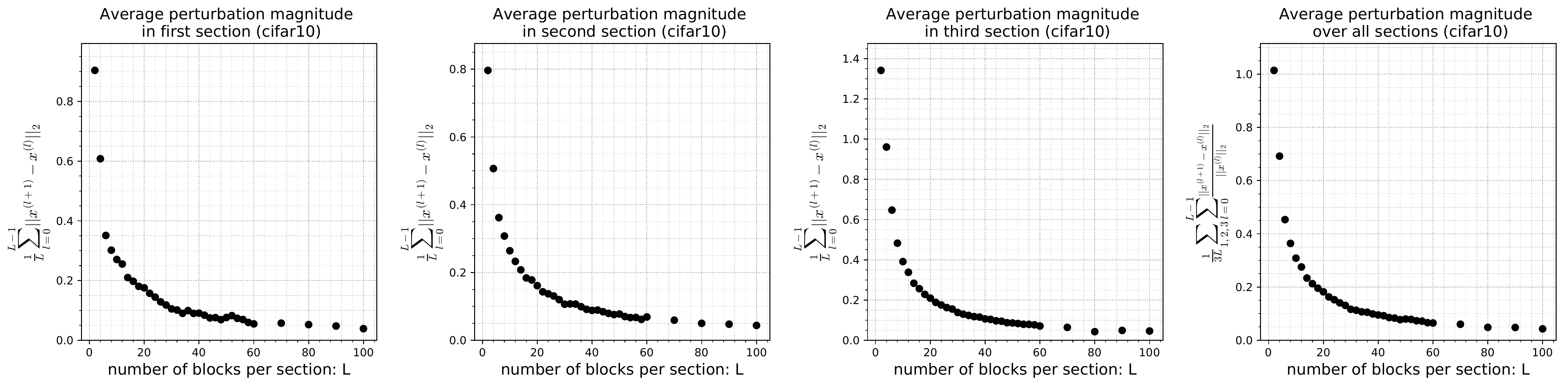}}
  \caption{Mean scaled perturbation magnitudes in the three residual sections of the network, namely $\frac{1}{L} \sum_{l'=0}^{L-1} \frac{||x^{(l+1)}-x^{(l)}||_2}{||x^{(l)}||_2} $ for $L=2,4,\dots,68,70,80,90,100$, as well as the mean perturbation magnitude over the entire network, namely $ \frac{1}{3\cdot L}\sum_{\text{section}=1,2,3} \sum_{l'=0}^{L-1} \frac{||x^{(l+1)}-x^{(l)}||_2}{||x^{(l)}||_2} $.}
\label{fig:avg-scaled-perturbation-size}
\end{figure*}

From Section~\ref{sec:measuring-computational-distance} we see a diminishing return on computational distance with increasing the total number of residual blocks. This suggests that the average size of the perturbation should decrease as the number of residual blocks per section increases, where the average size of the perturbation is given by $ \frac{1}{L} \sum_{l'=0}^{L-1} || x^{(l'+1)} - x^{(l')} ||_2 $. This is seen in Figure~\ref{fig:avg-distance-size}. We see that for both MNIST as well as in CIFAR10, in Figure~\ref{fig:avg-distance-size}a and Figure~\ref{fig:avg-distance-size}b respectively, the average magnitude of the transformation decreases as the number of residual blocks increases.

\subsection{Scaled Computational Distance}

As mentioned in Section~\ref{sec:math-section}, the computational distance $ || x^{(l+1)} - x^{(l)} ||_2 $ tells us the absolute magnitude of a single block's perturbation transformation. However, this is a difficult quantity to interpret since it is difficult to understand relative scales of what is a large or small transformation. With this motivation, we defined the \emph{scaled} computational perturbation as $ \frac{|| x^{(l+1)} - x^{(l)} ||_2}{|| x^{(l)} ||_2}  $. If this dimensionless variable is of order $1$ then we can say that $|| x^{(l+1)} - x^{(l)} ||_2$ is not a perturbation term, however if it is much less than $1$ then we can say that it can be interpreted as a perturbation term.

With this in mind we define the scaled computational distance as the sum of these perturbations $ \sum_{l'=0}^{L-1} \frac{|| x^{(l'+1)} - x^{(l')} ||_2}{|| x^{(l')} ||_2}  $, which can be seen in Figures~\ref{fig:scaled-comp-dist}a and~\ref{fig:scaled-comp-dist}b for both MNIST and CIFAR10, respectively. It is seen that in this scaled sense the same general characteristics of the absolute distance is maintained, in that for smaller networks (roughly $L<20$) there is a large increase in how much the image is transformed with a corresponding increase in the number of residual blocks per section. For medium sized networks (roughly $20<L<80$) there is a sharply diminished return on computational transformations with increase in the number of blocks per section. Finally for the large networks (roughly $L>80$) the computational capacity of the network seems to saturate and no further increase in computational transformations seem to take place.

\subsection{Average Scaled Perturbation Magnitude}
\label{sec:avg-scaled-per-mag}
We now turn our analysis towards beginning to understand, in the normalized scaled sense, how the magnitude of the perturbation changes with a corresponding increase in the number of blocks per residual section. This allows us to more readily understand if, and at what point, the residual network is infact learning a perturbation from identity. We do this by calculating Equation~\ref{eqn:delta-bar}, namely $ \frac{1}{L} \sum_{l'=0}^{L-1} \frac{|| x^{(l'+1)} - x^{(l')} ||_2}{|| x^{(l')} ||_2} $, which tells us the average magnitude of the scaled perturbations, with the results seen in Figures~\ref{fig:avg-scaled-perturbation-size}a and~\ref{fig:avg-scaled-perturbation-size}b for MNIST and CIFAR10, respectively.

If one says that the transformation constitutes a perturbation from identity if the relative scale is $0.1$, then on MNIST this is achieved at $40$, $32$ and $40$ residual blocks per layer for the first, second and third residual sections. Correspondingly on CIFAR10 this is achieved at $34$, $38$ and $48$ residual blocks per layer for the first, second and third residual sections. Note that on MNIST this totals to $112$ residual blocks, whereas on CIFAR10 this totals to $120$ residual blocks. This intuitively makes sense since CIFAR10 is a more complex dataset than MNIST and thus one would expect that it would take a larger network with more residual blocks to be able to create a small enough finite difference mesh in order to accurately the differential equation that transforms the data from low level input to high level output.

\subsubsection{Stopping Rule for Network Depth}

The calculated values for $\sum_{l'=0}^{L-1} \frac{|| x^{(l'+1)} - x^{(l')} ||_2}{|| x^{(l')} ||_2}$ for MNIST for the first, second and third residual sections are $d_{\text{scaled}}=4.11,3.12$ and $4.58$. This means that in order to have an $\epsilon$ perturbation be less than $0.1$, we should take $L=\text{ceil} \left( \frac{d_\text{scaled}}{\epsilon} \right)=37,39$ and $43$. Similarly for CIFAR10 the values of the scaled distance are $d_{\text{scaled}}=3.69,3.81$ and $4.27$, so in order to have $\epsilon<0.1$ we should take $L=\text{ceil} \left( \frac{d_\text{scaled}}{\epsilon} \right)=42,32$ and $36$. It is seen that these predicted values are fairly close to the actual experimental values found in Section~\ref{sec:avg-scaled-per-mag}, thus suggesting that this could be used as an automated way for defining the depth of residual networks.

\section{Conclusions}

The primary purpose of this paper is to test the hypothesis that residual networks are learning a perturbation from identity. This is an extremely important question since this hypothesis was a guiding principle in the development of residual networks, which, in their various forms, currently constitute the state-of-the-art in a huge variety of computer vision and machine learning tasks. In order to come to an answer, various measures were introduced to more concretely formulate, and therefore answer, this question. These measures were a computational distance and a magnitude of perturbation, as well as a scaled computational distance and scaled magnitude of perturbation, each of which have their relative merits and drawbacks. The ideas proposed in this paper were experimentally tested on both MNIST as well as CIFAR10, using residual networks with the total number of blocks ranging from $6$ to $300$. It was found that on MNIST and CIFAR10, and for sufficiently large residual networks, of the order $40$ residual blocks per section, the forcing function term is an order of magnitude less than the identity term and so the perturbation approximation is validated.


\end{document}